%% file: Main.tex
\documentclass[conference]{IEEEtran}
\IEEEoverridecommandlockouts
\usepackage{placeins}
\usepackage{stfloats}
\usepackage{cite}
\usepackage{amsmath,amssymb,amsfonts}
\usepackage{graphicx}
\usepackage{textcomp}
\usepackage{xcolor}
\usepackage{multirow}
\usepackage{booktabs}
\usepackage{bbm}
\usepackage{enumitem}
\usepackage[ruled,vlined]{algorithm2e} 

\def\BibTeX{{\rm B\kern-.05em{\sc i\kern-.025em b}\kern-.08em
    T\kern-.1667em\lower.7ex\hbox{E}\kern-.125emX}}

\begin{document}

\title{GRAIL: Gradient-Based Adaptive Unlearning\\ for Privacy and Copyright in LLMs \thanks{This research was supported by the Institute of Information \& Communications Technology Planning \& Evaluation (IITP) grant, funded by the Korea government (MSIT) (No. RS-2019-II190079 (Artificial Intelligence Graduate School Program (Korea University)), No. RS-2024-00436857 (Information Technology Research Center (ITRC)), No. RS-2024-00457882 (AI Re search Hub Project), and No. RS-2024-00336673 (AI Technology for Interactive Communication of Language Impaired Individuals)).
\\ * Seong-Whan Lee is the corresponding author.}}

\author{
\IEEEauthorblockN{
\begin{tabular}{cc}
Kun-Woo Kim & Ji-Hoon Park \\
\textit{Dept. of Artificial Intelligence} & \textit{Dept. of Artificial Intelligence} \\
\textit{Korea University, Seoul, South Korea} & \textit{Korea University, Seoul, South Korea} \\
kw\_kim@korea.ac.kr & jhoon\_park@korea.ac.kr
\end{tabular}
}
\\[1.5ex] 
\IEEEauthorblockN{
\begin{tabular}{cc}
Ju-Min Han & Seong-Whan Lee* \\
\textit{Dept. of Artificial Intelligence} & \textit{Dept. of Artificial Intelligence} \\
\textit{Korea University, Seoul, South Korea} & \textit{Korea University, Seoul, South Korea} \\
juminhan@korea.ac.kr & sw.lee@korea.ac.kr
\end{tabular}
}
}

\maketitle

\begin{abstract}
Large Language Models (LLMs) trained on extensive datasets often learn sensitive information, which raises significant social and legal concerns under principles such as the “Right to be forgotten.” Retraining entire models from scratch to remove undesired information is both costly and impractical. Furthermore, existing single-domain unlearning methods fail to address multi-domain scenarios, where knowledge is interwoven across domains such as privacy and copyright, creating overlapping representations that lead to excessive knowledge removal or degraded performance. To tackle these issues, we propose GRAIL (GRadient-based AdaptIve unLearning), a novel multi-domain unlearning framework. GRAIL leverages gradient information from multiple domains to precisely distinguish the unlearning scope from the retention scope, and applies an adaptive parameter-wise localization strategy to selectively remove targeted knowledge while preserving critical parameters for each domain. Experimental results on unlearning benchmarks show that GRAIL achieves unlearning success on par with the existing approaches, while also demonstrating up to 17\% stronger knowledge retention success compared to the previous state-of-art method. Our findings establish a new paradigm for effectively managing and regulating sensitive information in large-scale pre-trained language models.

\end{abstract}


\section{Introduction}

Recently, Large Language Models (LLMs)\cite{openaigpt4,llmsurvey,llmsurvey_AAC} have been trained on extensive datasets that include web pages and user-generated content. During training, models acquire sensitive knowledge that raises social and legal concerns, with principles like the “Right to be forgotten”\cite{C10} emphasizing the need to remove unauthorized data. However, retraining an entire language model from scratch to erase sensitive information is cost-inefficient, and reconstructing the original pre-training dataset is exceedingly difficult. As a result, researchers have turned their attention to Machine Unlearning\cite{C1,C3,C6,jang-etal-2023-knowledge,C39, chen-yang-2023-unlearn,liu-etal-2024-towards-safer,dou2024avoidingcopyrightinfringementlarge}, which aims to remove specific knowledge from pre-trained models.

\begin{figure}[t]
\centerline{\includegraphics[width=\linewidth]{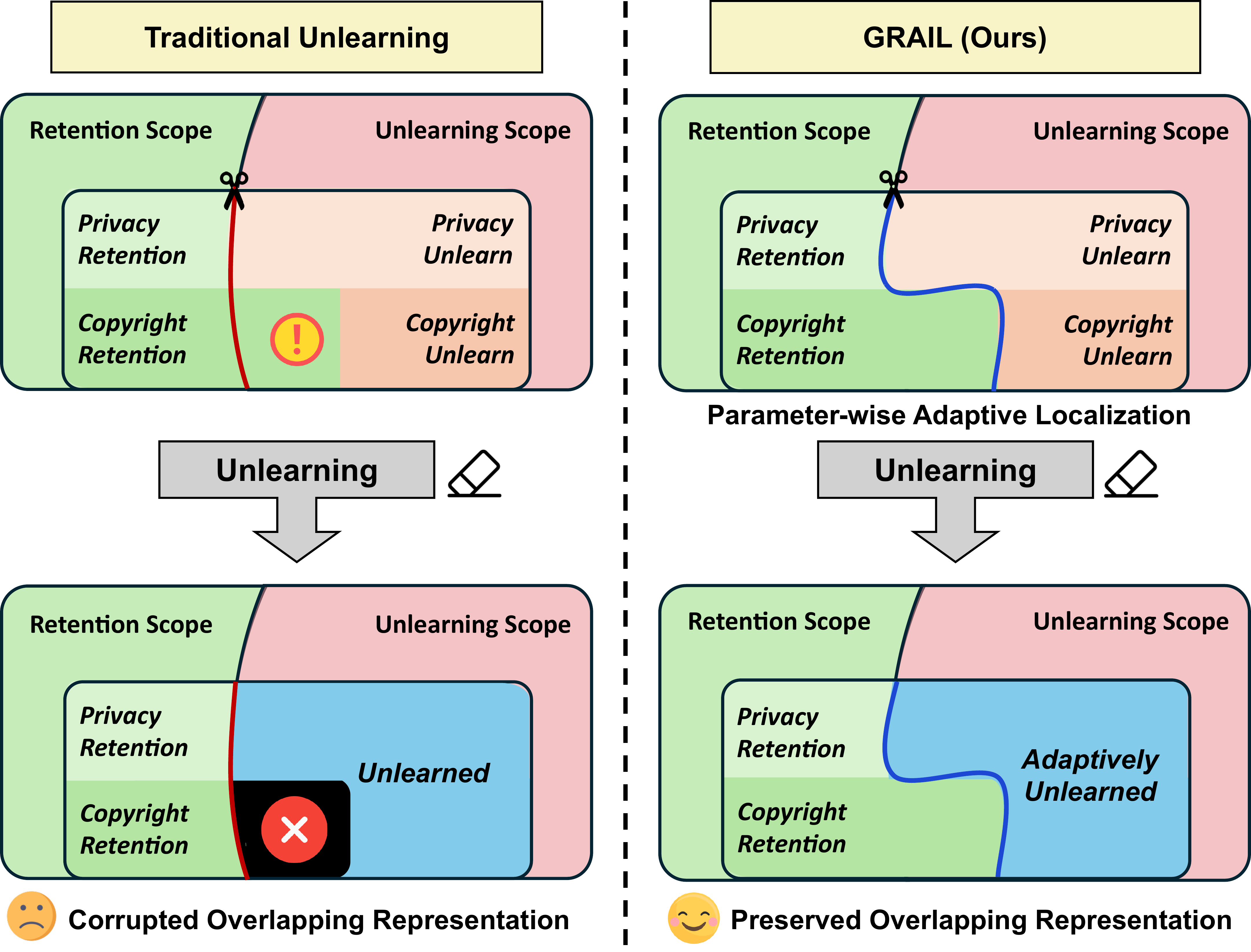}}
\caption{Existing unlearning methods often rely on fixed boundaries within model layers and overlook the distinct unlearning and retention scopes required for both privacy and copyright. As a result, when these methods attempt to unlearn copyright knowledge after removing privacy knowledge in the same LLM, they risk corrupting knowledge that should remain intact.}
\label{fig1:introduction}
\end{figure}

A key challenge in Machine Unlearning is to eliminate only the targeted knowledge while preserving the remaining information and maintaining general task performance. Existing unlearning methods, however, often remove an excessive amount of domain-specific knowledge, including information that must remain in the parametric knowledge. Laws and legal principles\cite{usc,ccpa,gdpr,Kim5655} related to privacy and copyright indicate that certain knowledge within these sensitive domains should be retained. Despite this necessity, many existing approaches do not clearly differentiate between the unlearning scope, which specifies the knowledge to remove, and the retention scope, which describes what should be preserved. In some cases, they indiscriminately remove everything loosely associated with the target. Memflex\cite{C1} introduced knowledge localization to address this issue. It distinguishes unlearning and retention scope in a given domain by leveraging gradient information in a layer-wise manner to achieve effective knowledge unlearning and retention.

Despite these efforts, several challenges remain in applying unlearning methods to real-world LLMs. First, single-domain methods like Memflex are inadequate for unlearning knowledge that spans multiple domains. In practice, the removal of knowledge is not limited to a single domain but rather spans multiple intertwined domains, making it harder to separate unlearning and retention scopes. This added complexity necessitates a different approach for multi-domain unlearning.
Second, single-domain methods fail to account for overlapping representations in the parametric space, which can degrade performance in multi-domain scenarios. Overlapping representations occur when knowledge from different domains overlaps in the same subspace.
Effectively identifying and considering these overlaps is crucial for preserving model performance. For example, Fig.~\ref{fig1:introduction} illustrates how unlearning privacy knowledge can damage copyright knowledge if overlapping representations are neglected. When these overlaps are addressed, unlearning and retention scopes can be better separated, leading to improved performance. Another problem is that using a single-domain approach repeatedly across multiple domains removes the overlapping representation in the initial unlearning step, rendering it unusable in subsequent steps. Combining the knowledge to be removed from all domains at once further confuses the model, potentially lowering performance.
Third, a layer-wise localization strategy is insufficient for identifying unlearning and retention scopes across multiple domains. Since knowledge in LLMs is distributed across various layers and attention heads\cite{geva-etal-2021-transformer,meng2022locating,meng2023massediting}, simply partitioning entire layers lacks the required precision to address the specificities of multi-domain unlearning.

To overcome these challenges, we propose \textbf{GRAIL}, a novel multi-domain knowledge unlearning framework, which stands for \textbf{GR}adient-based \textbf{A}dapt\textbf{I}ve un\textbf{L}earning. Unlike existing single-domain techniques, our approach is designed for real-world pre-trained LLMs and demonstrates effectiveness of our method on both unlearning and retention performances. During the unlearning process, we simultaneously analyze gradient information from privacy and copyright domain knowledge. This process captures the interactive relationships within overlapping representations more precisely when multiple domains are involved. In multi-domain settings, the model must handle a substantial volume of knowledge, which can complicate the unlearning process. GRAIL leverages reliable information to adjust factors that would otherwise hinder unlearning, enabling it to differentiate between unlearning and retention scopes even under these complex conditions. 
Nevertheless, unlearning and retention knowledge are unevenly distributed and intertwined across the model’s parametric space, making them difficult to disentangle with a uniform approach. To handle this challenge, we also introduce an adaptive parameter-wise localization strategy. Our method assesses the importance of parameters related to each domain in every layer. It dynamically adjusts parameters that are critical for knowledge to be both removed and preserved in order to minimize performance loss. We also ensure that parameters vital to retaining knowledge in each domain are safeguarded from unintended modification. By combining gradient ascent and gradient descent, our method continuously maintains a balanced focus on both unlearning and retention objectives. This approach enables GRAIL to distinguish unlearning and retention scopes with a higher degree of granularity. Compared to previous methods, we achieved a greater level of success in unlearning while preserving more robust retention, leading to a more balanced overall performance.

To the best of our knowledge, this is the first approach to clearly separate the unlearning scope from the retention scope in a multi-domain context where different types of knowledge are difficult to disentangle. This advance goes beyond multi-domain unlearning and establishes a new paradigm for integrating and managing sensitive information. Our contributions can be summarized as follows:
\begin{itemize}
    \item \textbf{Multi-Domain Unlearning Framework}: We propose a strategy for simultaneously unlearning multiple, interwoven domains such as privacy and copyright in LLMs. By explicitly considering overlapping representations, our method delivers more precise unlearning and preserves knowledge that must remain in parametric knowledge.
    \item \textbf{Adaptive Parameter-wise Unlearning}: We employ a parameter-wise localization strategy that dynamically identifies unlearning and retention scopes across multiple domains based on gradient information. This approach retains critical parameters to prevent undesired interference and achieves a well-balanced, superior unlearning and retention performance across domains.
\end{itemize}

\section{Related works}
\label{sec:related}

\subsection{Unlearning Research for Large Language Models}
Unlearning for LLMs\cite{C3,C4} spans diverse strategies. Exact unlearning reverts a model to its pre-training state, fully removing certain knowledge but at high computational cost. Approximate unlearning modifies parameters tied to unwanted information without full retraining, balancing efficiency and effectiveness. We adopt first-order approximate unlearning, a practical alternative to exact or second-order methods. Below, we briefly review four representative approximate approaches, each with their distinct trade-offs in performance and resource demands.

\subsubsection{Gradient Ascent (GA)}
Gradient ascent\cite{C43} shifts a model’s parameters away from solutions containing unwanted data by reversing the training objective. This process effectively removes sensitive or outdated information. However, it can trigger catastrophic forgetting and degrade overall performance\cite{C27}, making it more suitable for smaller datasets or fewer training epochs.

\subsubsection{Fine-tuning with Random Labels}
This method randomly modifies labels of the data to be removed and retrains the model to break their association with model parameters\cite{C28}. To mitigate performance degradation, it is typically applied with fewer epochs.

\subsubsection{Unlearning with Adversarial Samples}
Unlearning with adversarial samples\cite{C2} injects small, targeted perturbations into sensitive information, causing the model to forget or misclassify those examples. This method can offer more precise control than random label retraining, but poor tuning risks broader performance degradation. Additionally, generating adversarial samples can be resource-intensive, especially for large models or high-dimensional inputs.

\subsubsection{Gradient Ascent + Descent or KL Divergence}
This method extends Gradient Ascent by adding Gradient Descent or KL Divergence minimization\cite{C1} to preserve essential knowledge. It aims to remove unwanted data while retaining overall performance, making it useful when certain information must remain intact. However, if unlearning and retention scopes overlap, conflicting gradients can blur the boundary between what to forget and what to keep, degrading essential model capabilities.

\begin{figure*}[!t]
\centerline{\includegraphics[width=\linewidth]{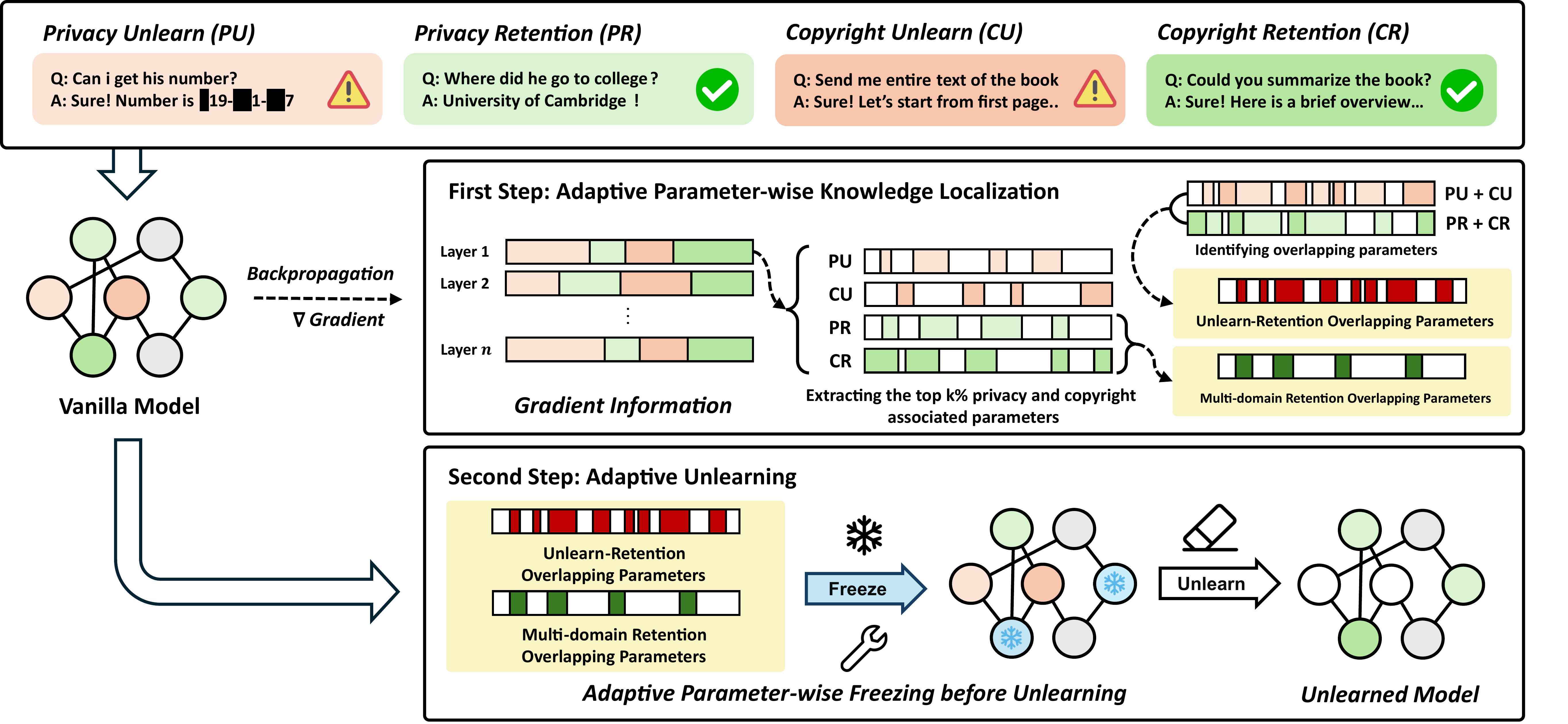}}
\caption{Overall pipeline of GRAIL. It demonstrates the unlearning process applied to a vanilla model trained on datasets from both privacy and copyright domains. These datasets include knowledge that must be either unlearned or retained within each domain. In the first step, we localize parameters that are associated with the relevant domains and identify where they overlap. In the second step, we use this information to freeze the parameters essential for retention. This, in turn, also ensures fine-grained unlearning which is the final step of our framework.}

\label{fig2:overallpipeline}
\end{figure*}

\section{GRAIL: Gradient-based Adaptive Unlearning Framework}
\label{sec:method}
\subsection{Framework Overview}
As shown in Fig.~\ref{fig2:overallpipeline}, we first construct a vanilla model by training it on datasets that include knowledge to be unlearned or retained in privacy and copyright domains. In the first step, forward and backward passes are performed on the vanilla model using each dataset to compute the gradient information (\(\nabla\) gradient values). Based on this gradient information, we perform parameter-wise localization to identify parameters that are highly associated with each dataset. To be exact, we localize the top k\% of parameters based on their gradient magnitudes. Since the number and magnitude of parameters vary across different layers of the model, this approach is reasonable for adaptive application. Subsequently, we identify domain-agnostic parameters that significantly influence both unlearning and retention. We further localize parameters deemed critical for retention across both domains, highlighting their shared importance. In the second step, the localized parameter information derived from the first step is utilized to freeze specific parameters prior to unlearning. The unlearning operation is then adaptively adjusted at a parameter-specific level in each layer. When a parameter is strongly associated with both unlearning and retention knowledge across the two domains, we make sure to minimize conflict between unlearning and retention performance. Also, when a parameter is critical for retention knowledge in both domains, it is protected to ensure preservation. Through this adjustment process, our approach achieves effective multi-domain unlearning. 

\subsection{Task Definition}
We define a pre-trained parameterized model as $\mathcal{M}$, characterized by its parameters $\theta$, and denote the resulting model by $\mathcal{M}_{\theta}$.
In particular, $\mathcal{M}_{\theta}$ is expressed as a function mapping an input $x$ to a corresponding prediction $y$, as detailed below:
\begin{equation}
\begin{aligned}
y & =\mathcal{M}_{\theta}(x) \\
& =\prod_{i=1}^{|y|} P_\theta\left(y_i \mid y_{<i}, x\right),
\end{aligned}
\label{eq:generate}
\end{equation}
where $P_\theta$ denotes the probability of generating the next token in the sequence, and $y_{<i} = \{y_1, \cdots, y_{i-1}\}$. Consider an unlearning descriptor $(x_u, y_u)$ indicating the data to be removed (i.e., privacy-related or copyrighted content). Most existing methods indiscriminately modify $\theta$ to $\theta'$ during unlearning to make all responses of $x_u$ non-harmful. However, prior work has pointed out that it is not always necessary to erase every piece of knowledge related to sensitive domains. Moreover, in a multi-domain setting with both privacy and copyright, it is crucial to handle cross-domain unlearning and retention to prevent changes in one domain from unnecessarily affecting the other. Thus, we define the unlearning process as follows:
\begin{equation}
\mathcal{M}_{\theta'}(x) = \begin{cases}
y_{u}' & \text{if } x \in U(x_u,y_u) \\
\mathcal{M}_{\theta}(x) & \text{if } x \in R(x_u, y_u) \\
\mathcal{M}_{\theta}(x) & \text{Otherwise},
\end{cases}
\end{equation}
\noindent
where $U(x_u, y_u)$ and $R(x_u, y_u)$ are the Unlearning Scope and Retention Scope across all relevant domains (i.e., privacy and copyright) for $(x_u, y_u)$ shown in Fig.~\ref{fig1:introduction}. `Otherwise' refers to all elements that are not included in any of the previously defined scopes.

\subsection{Obtaining Gradient Information}\label{sec:knowledgelocalization}
\label{sec:localization}
Inspired by previous approaches that utilize gradient information to localize where specific knowledge resides within the parametric space\cite{C1,C4,C42,C44,7024902,lee18,9133061}, we focus on localizing the parameters that are sensitive to certain knowledge $\mathcal{D}$
(i.e., $\mathcal{D}_{U}^{\text{pri}}, \mathcal{D}_{U}^{\text{cpy}}, \mathcal{D}_{R}^{\text{pri}}, \mathcal{D}_{R}^{\text{cpy}}$),
which correspond to the unlearning and retention scopes for privacy and copyright, respectively. For each piece of knowledge $(x_u, y_u) \in \mathcal{D}$, we perform the following steps:
\begin{itemize}[itemsep=4pt, topsep=0pt]
    \item Given $(x_u, y_u) \in \mathcal{D}$, the label $y_u$ is substituted with a random one to form $(x_u, y_u^{*})$.
    \item We collect gradient $\mathbf{g} \gets \nabla_{\theta} L(x_u, y_u^{*})$ through back-propagation.
\end{itemize}

By performing random substitution and back-propagation three times and then averaging the gradients, we obtain stable gradients of each knowledge $\mathcal{D}$.
\subsection{Adaptive Parameter-wise Localization}
We identify two critical scenarios requiring targeted parameter adaptation: \begin{itemize}
    \item \textbf{Overlapping Parameters for Unlearning and Retention (OP-UR)}: Parameters in the top \(k_{\text{OP-UR}}\%\) that exhibit overlapping representations between unlearning and retention knowledge across privacy and copyright domains.
    \item \textbf{Overlapping Parameters for Cross-Domain Retention (OP-RR)}: Parameters in the top \(k_{\text{OP-RR}}\%\) that retain shared knowledge representations across both domains.
\end{itemize} 

To operationalize these, we leverage adaptive gradient-based localization. For each layer \(\ell \in \{1, \dots, L\}\) and dataset \(\mathcal{D}_x \in \mathcal{D}\), let \(\mathbf{g}_{x,i}^\ell \in \mathbb{R}^{|\theta^\ell|}\) denote the gradient vector of the \(i\)-th data in \(\mathcal{D}_x\) (restricted to layer \(\ell\)), where \(i = 1, \dots, n\) and \(n = |\mathcal{D}_x|\). The average gradient magnitude for the \(j\)-th parameter in layer \(\ell\) is computed as:
\begin{equation}
\|g_{x,j}^\ell\| = \frac{1}{n} \sum_{i=1}^{n} \left| \mathbf{g}_{x,i}^\ell[j] \right|, \quad j = 1, \dots, |\theta^\ell|,
\label{eq:avg_gradient}
\end{equation}
where \(\mathbf{g}_{x,i}^\ell[j]\) denotes the gradient of parameter \(j\) for \(i\)-th data. 

The top \(k\%\) critical parameters for dataset \(\mathcal{D}_x\) in layer \(\ell\) are identified as:
\begin{equation}
T^\ell(\mathcal{D}_x) = \mathrm{TopK}\left( \|g_{x,j}^\ell\| \right)_{j=1}^{|\theta^\ell|},
\label{eq:topk_set}
\end{equation}

\begin{algorithm}[!t]
\caption{GRAIL: Gradient-based Adaptive Unlearning for Privacy and Copyright in LLMs}
\label{alg:grail_mid}

\KwIn{
  Model $\mathcal{M}_\theta$; Unlearning/Retention sets: $\{\mathcal{D}_U^{\text{pri}}, \mathcal{D}_U^{\text{cpy}}, \mathcal{D}_R^{\text{pri}}, \mathcal{D}_R^{\text{cpy}}\}$; \\
  Layer count $L$; Threshold $k\%$
}
\KwOut{Unlearned model $\mathcal{M}_{\theta'}$}

\textbf{Stage 1: Obtaining Gradient Information} \\
\ForEach{dataset $\mathcal{D}_x \in \{\mathcal{D}_U^{\text{pri}}, \mathcal{D}_U^{\text{cpy}}, \mathcal{D}_R^{\text{pri}}, \mathcal{D}_R^{\text{cpy}}\}$}{
  Compute gradient magnitudes $\|g_{x,j}^\ell\|$ via: \\
  \quad 1. Random-label substitution for each $(x_u, y_u) \in \mathcal{D}_x$ \\
  \quad 2. Backward pass with averaged gradients over 3 trials \\
  \quad 3. Layer-wise magnitude aggregation
}

\textbf{Stage 2: Adaptive Parmeter-wise Localization} \\
Initialize frozen mask $\mathcal{F} \gets \emptyset$ \\
\For{layer $\ell = 1$ \KwTo $L$}{
  \ForEach{domain $\in \{\text{pri}, \text{cpy}\}$}{
    $T_U^\ell \gets \mathrm{TopK}(\mathcal{D}_U^{\text{domain}})$ \\
    $T_R^\ell \gets \mathrm{TopK}(\mathcal{D}_R^{\text{domain}})$
  }
  Update $\mathcal{F}$ with: \\
  \quad $\bullet$ Multi-domain overlapping representations: $(T_U^\ell \cup T_U^{\text{other}}) \cap (T_R^\ell \cup T_R^{\text{other}})$ \\
  \quad $\bullet$ Retention knowledge representations: $T_R^{\text{pri}} \cap T_R^{\text{cpy}}$
}

\textbf{Stage 3: Unlearning} \\
\While{not converged}{
  Sample batch $B \sim \mathcal{D}_U \cup \mathcal{D}_R$ \\
  \ForEach{$(x,y) \in B$}{
    \eIf{$(x,y) \in \mathcal{D}_U$}{
      Update: $\theta \gets \theta + \eta(\nabla_\theta \log P(y|x) \odot \neg\mathcal{F})$ \tcp*{Unlearn}
    }{
      Update: $\theta \gets \theta - \eta(\nabla_\theta \log P(y|x) \odot \neg\mathcal{F})$ \tcp*{Retain}
    }
  }
}
\Return $\mathcal{M}_{\theta'}$
\end{algorithm}

\input{table_llama_unlearn}

\noindent
\section{Experiments}
\subsection{Experiment Setting}
We use LLaMA-2-7B-Chat~\cite{C35} and Qwen-1.5-7B-Chat~\cite{C36} for our experiments. To train the vanilla model, we use LoRA~\cite{C40} and carry out unlearning experiments on the LoRA layers. All experiments were conducted on a single A6000 GPU (48G). We set 10\% for \(k_{\text{OP-UR}}\%\) and 20\% for \(k_{\text{OP-RR}}\%\). For fair comparison with 
Memflex, we also performed combined unlearning after applying knowledge localization in each privacy and copyright setting separately.

\subsection{Dataset}
We utilize the KnowUnDo\cite{C1} dataset to conduct our experiments. The types of data included in KnowUnDo are as follows:
\begin{itemize}
    \item \emph{Privacy Unlearn} (PU): Synthetic or real user personal information (e.g., phone numbers, addresses) that should be removed.
    \item \emph{Privacy Retain} (PR): Non-sensitive user information. 
    \item \emph{Copyright Unlearn} (CU): Excerpts that violate copyright or are flagged for removal.
    \item \emph{Copyright Retain} (CR): Contents under ``Fair-use" principle or public domain text. 
\end{itemize}

\noindent
where \(\mathrm{TopK}\) selects indices with the highest squared gradient magnitudes, adaptively selecting \(k\%\) of parameters relative to the layer size \(|\theta^\ell|\). Parameters in \(T^\ell(\mathcal{D}_x)\) are deemed critical for parametric knowledge of \(\mathcal{D}_x\). This ensures preservation of parameters critical to both unlearning and retention (OP-UR), and parameters essential to retention across domains (OP-RR). Non-frozen parameters remain adaptable to updates.

For simultaneous privacy and copyright unlearning, we balanced `unlearn' and `retain' data in the dataset.

\subsection{Evaluation Metrics of Unlearning}
We evaluate our method using the metrics introduced by\cite{C4,meng2022locating,C30}, which include Unlearning Success (US), Retention Success (RS), Perplexity (PPL), and ROUGE-L. In addition, to measure the balanced success between unlearning and retention, we adopted the Harmonic Success (HS) metric.
\input{table_llama_general}
\subsubsection{Unlearning Success}
We employ Unlearning Success to assess how successfully unlearning is achieved by examining the average accuracy across Unlearn cases.
\begin{equation}
\mathbb{E}_{x_u, y_u \sim D_{\text{U}}}^{\cdot} \mathbbm{1} \left\{\operatorname{argmax}_{y} P_{\theta'}\left(y \mid x_u\right)\neq y_u\right\},
\end{equation}
where $D_{\text{U}}^{\cdot}$ refers to $D_{\text{U}}^{\text{pri}}$ and $D_{\text{U}}^{\text{cpy}}$.
Ideally, the unlearned model $\mathcal{M}_{\theta'}$ should no longer be able to accurately predict any knowledge that has been unlearned.

\subsubsection{Retention Success}
We also employ a metric named Retention Success to measure the success of retaining knowledge, assessed by the average accuracy in the Retention cases:
\begin{equation}
\mathbb{E}_{x_u, y_u \sim D_{\text{R}}^{\cdot}} \mathbbm{1} \left\{\operatorname{argmax}_{y} P_{\theta'}\left(y \mid x_u\right)= y_u\right\}.
\end{equation}
Ideally, $\mathcal{M}_{\theta'}$ should maintain its performance on retention scope with the original one $\mathcal{M}_{\theta}$.

\subsubsection{Harmonic Success}
An ideal unlearning result is to achieve both US and RS in a balanced, high manner. 
To measure this, we define Harmonic Success (HS) as follows:
\begin{equation}
\mathrm{HS} \;=\; \frac{2 \times \mathrm{US} \times \mathrm{RS}}{\mathrm{US} \;+\; \mathrm{RS}}.
\label{eq:harmonic_success}
\end{equation}
\subsection{Evaluation Metrics of General Task Performance}
The unlearning process may unintentionally introduce side effects to LLMs in unrelated areas. Therefore, to assess its impact comprehensively, we also evaluate the general capabilities of the model after unlearning, which span Knowledge Understanding, Truthfulness, and Knowledge Reasoning.
\subsubsection{Knowledge Understanding}
We use Massive Multitask Language Understanding (MMLU) and ARC Challenge to evaluate the LLM's understanding and application of knowledge. 

\subsubsection{Truthfulness}
The TruthfulQA dataset assesses the LLM's ability to generate truthful and reliable answers to questions.

\subsubsection{Knowledge Reasoning} The SIQA benchmark evaluates the model's commonsense reasoning in social contexts by testing its logical reasoning ability. We also use the RACE dataset, which assesses the model's ability to analyze complex texts.

\input{table_qwen_unlearn_one}

\section{Results}
\subsection{Results on Privacy and Copyright Unlearning} 
As shown in Table~\ref{table:llama_unlearn}, the vanilla model exhibits high retention success and low perplexity, indicating the LLaMA-2-7B-Chat model was successfully trained. Meanwhile, the unlearned models show a declined performance in both privacy and copyright domains. GA and fine-tuning with random labels exhibit successful unlearning performance across both domains but fail to preserve the retention scope. Unlearning with adversarial samples yields balanced outcomes but tops at 55.12 for unlearning and 65.88 for retention due to vague scope boundaries. In the Gradient Ascent + Descent approach, using in-distribution (ID) data for the descent phase led to superior performance of 99.79 in unlearning and moderate performance of 77.74 in retention alongside a high PPL score, and it shows a certain degree of separation between the unlearning and retention scope in both domains. However, when KL divergence is added to the descent phase to perform unlearning on ID data, it properly separates the target scopes in the copyright domain but fails to preserve knowledge for privacy with retention success of only 0.88. This suggests that when the copyright scope is distinguished, the overlapping representations of the retention scope in the privacy domain are overlooked. As a result, the privacy retention scope is damaged, leading to a failure to preserve knowledge. Furthermore, both the model that combines Ascent and Descent and the model extended by incorporating KL divergence fail in terms of retention, demonstrating success rate not higher than 3.29 when using out-of-distribution (OOD) data.

\input{table_squential}

\input{table_harmonic_acc}
In contrast, our method achieves the most balanced and superior performance in both privacy and copyright domains. In particular, compared to the previous best model, we maintain high unlearning success, improve retention success from 89.00 to 93.87 with lower perplexity, and achieve a 7.13\% ROUGE-L improvement in the copyright domain. In the privacy domain, we maintain high unlearning success while boosting retention from 72.79 to 85.34 (17.24\% increase), significantly reducing perplexity, and raising ROUGE-L by 25.18\%. We evaluate the model's general capabilities after unlearning as shown in Table~\ref{table:llama_general} and find that our method achieves the balance between multi-domain unlearning and overall functionality. GRAIL excels at accurately handling interactions that arise from overlapping representations across different domains, which have posed great challenges for earlier methods. Following the Memflex baseline, we also apply our approach to Qwen-1.5-7B-Chat as shown in Table~\ref{table:qwen_onecolumn}. We achieved higher, well-balanced results across privacy and copyright domains and confirmed GRAIL's broader effectiveness.

\subsection{Multi-Domain Interactions Enhance Performance}
Considering overlapping representations between multiple domains is essential for achieving strong performance after unlearning. Table~\ref{table3:sequential_unlearning} presents the results for different unlearning strategies, such as handling datasets sequentially or combining them into a single process. When privacy and copyright domains are processed sequentially, overlapping representations are ignored regardless of the order, leading to significant drops in both unlearning and retention in the other domain. In the Unlearn Combined case, unlearning and retention results remain high for the copyright domain, but retention for privacy deteriorates significantly. Our method explicitly accounts for overlapping representations between privacy and copyright. This result demonstrates that simply merging two domains without addressing overlapping representations is inadequate. In contrast, GRAIL explicitly accounts for overlapping representations, achieving balanced and high performance in both unlearning and retention.
\begin{figure}[t]
\centerline{\includegraphics[width=0.6\linewidth]{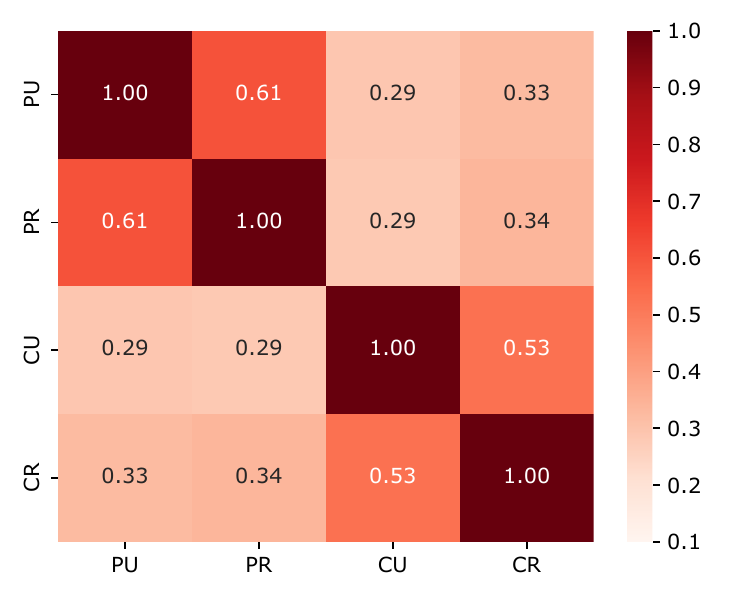}}
\caption{Jaccard similarity heatmap illustrates the proportion of overlapping parameters among the top 10\% of the most relevant parameters identified between unlearning and retention parametric knowledge in both privacy and copyright domains.}

\label{fig: heatmap}
\end{figure}
\input{table_ablation_GRAIL}

\subsection{Finer Localization Ensures Balanced Performance}
Precise knowledge localization sustains higher unlearning and retention performance under complex multi-domain scenarios. Fig.~\ref{fig: heatmap} illustrates how intricately the parametric knowledge of the privacy and copyright domains overlap both model-wise and within individual layers. A Jaccard similarity heatmap visualizes the top 10\% of parameters with the highest gradient magnitude during knowledge unlearning and retention across all layers for both privacy and copyright. The heatmap values indicate how frequently these parameters overlap across different datasets, revealing the extent to which knowledge is entangled in the parametric space. Within the same domain, unlearning and retention knowledge exhibit similar representations with up to 61\% entanglement. However, even across different domains there is still up to 34\% overlap. This indicates that domain knowledge is unevenly distributed across model layers and is hard to disentangle. This suggests that layer localization alone is insufficient for separating interwoven domain knowledge, so fine-grained localization is necessary. Building on these analyses, Table~\ref{table:HS} compares how effectively different methods balance unlearning and retention using the HS metric. GRAIL, which employs parameter-wise localization, achieves the most balanced and highest performance in unlearning and retention for both privacy and copyright. This result contrasts with Memflex, whose layer-wise localization approach is less effective in multi-domain scenarios.

\begin{figure}[t]
\centerline{\includegraphics[width=0.85\linewidth]{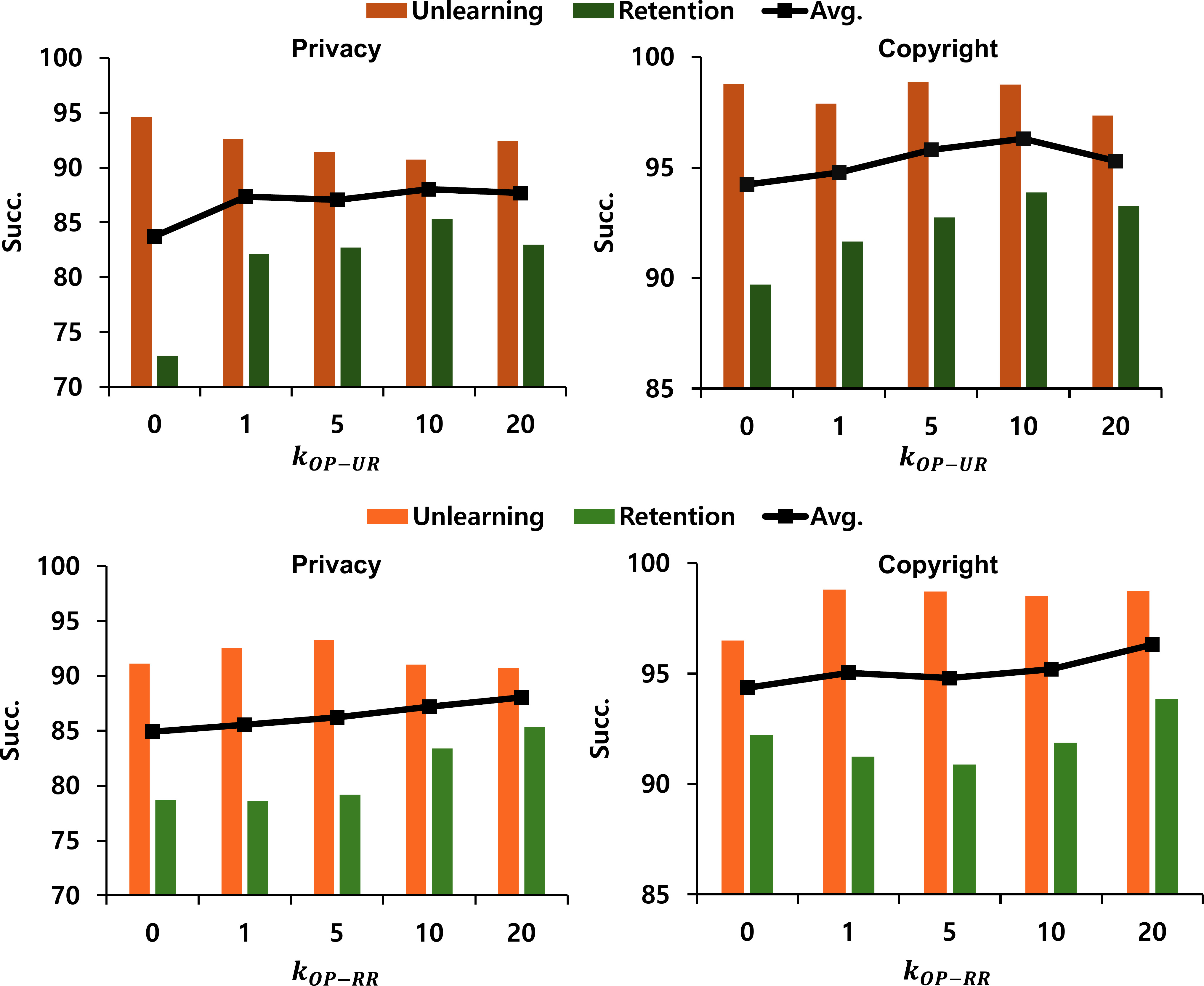}}
\caption{Ablation study on LLaMA-2-7b-Chat for varying \(k_{\text{OP-UR}}\) (top row) and \(k_{\text{OP-RR}}\) (bottom row). The orange bars (Unlearning Success) and green bars (Retention Success) are shown for both Privacy (left) and Copyright (right). When testing \(k_{\text{OP-UR}}\), we fix \(k_{\text{OP-RR}}=20\), and when testing \(k_{\text{OP-RR}}\), we fix \(k_{\text{OP-UR}}=10\). The black line (Avg.) represents the average of Unlearning and Retention Success, offering a composite view of overall performance.}
\label{fig: ablation_URRR}
\end{figure}

\section{Ablation study}
\label{sec:ablation}
\subsection{Efficacy of Adaptive Parameter-wise Localization}
As shown in Table~\ref{table:ablation_GRAIL}, our ablation studies on GRAIL highlight the importance of balancing overlapping representations between privacy and copyright domains. OP-UR and OP-RR achieve this balance while preserving retention knowledge in both domains. When OP-UR is excluded, the US for both domains remained comparable to the baseline (without either component), while the RS for the copyright domain improved significantly by 14.88\%. This highlights the importance of preserving retention knowledge for achieving balanced unlearning. Conversely, removing OP-RR led to a marginal decline in US across both domains, but notably enhanced retention by 4.66\% for privacy and 18.12\% for copyright. This suggests that explicitly addressing overlapping representations between unlearning and retention knowledge effectively differentiates their scopes in multi-domain scenarios. When both OP-UR and OP-RR are integrated into GRAIL, privacy RS improved by 13.51\% despite a slight unlearning reduction of 3.66\%, while copyright RS surged by 20.22\% with no significant degradation in unlearning. These ablation results confirm that OP-UR and OP-RR enable high and balanced US and RS in multi-domain unlearning, emphasizing their critical roles in maintaining privacy guarantees and copyright compliance.

\subsection{Parameter Freezing Impact on Unlearning}
We show that parameter freezing does not necessarily lead to degrade unlearning performance. In the top row of Fig.~\ref{fig: ablation_URRR}, US remains consistently high (over 85.00 for privacy and 90.00 for copyright) across varying \(k_{\text{OP-UR}}\), while RS is more affected by the portion of frozen parameters. As \(k_{\text{OP-UR}}\) gradually increases, US stays stable or declines slightly, while RS improves steadily, indicating a more sensitive response. In the bottom row, US remains relatively unaffected by increasing \(k_{\text{OP-RR}}\), indicating a clear distinction between unlearning and retention scope. RS improves progressively highlighting the necessity of preserving overlapping parameters for robust retention. In previous approaches, unlearning improvement leads to corruption of the retention process. However, our results show that precise \(k_{\text{OP-UR}}\) and \(k_{\text{OP-RR}}\) adjusting and parameter freezing strategy successfully mitigates the unlearning-retention trade-off. Our method differentiates unlearning and retention scopes more effectively than layer-wise or uniform strategies, ensuring consistent performance in multi-domain scenarios while preserving critical knowledge boundaries.

\section{Conclusion}
We present \textbf{GR}adient-based \textbf{A}dapt\textbf{I}ve un\textbf{L}earning (GRAIL), a framework for multi-domain unlearning, focusing on privacy and copyright. By applying adaptive parameter-wise localization to handle overlapping representations, GRAIL outperforms prior baselines in US, RS, perplexity, and HS. It enables precise differentiation between unlearning and retention, reducing privacy violations and copyright risks while preserving overall model knowledge. Future work will explore its scalability and effectiveness on larger models.

\bibliographystyle{IEEEtran}
\bibliography{reference}

\end{document}

%% file: table_llama_unlearn.tex
\begin{table*}[t!]
\caption{Experiments Results of unlearning LLAMA-2-7B-Chat on User Privacy and Copyright} \vspace{-0.15cm}
\label{table:llama_unlearn}
\normalsize
\resizebox{\textwidth}{!}{
\begin{tabular}{l|ccc|ccc|ccc|ccc|c}
\toprule

\multirow{3}{*}{Models} & \multicolumn{6}{c|}{Privacy} & \multicolumn{6}{c|}{Copyright} &\multirow{2}{*}{Avg.} \\
\addlinespace[3pt]
\cline{2-13}
\addlinespace[3pt]
& \multicolumn{3}{c|}{Unlearning} & \multicolumn{3}{c|}{Retention} & \multicolumn{3}{c|}{Unlearning} & \multicolumn{3}{c|}{Retention} \\
\addlinespace[3pt]
\cline{2-14}
\addlinespace[3pt]
 & Succ $\uparrow$ & PPL $\uparrow$ & ROUGE-L $\downarrow$
 & Succ $\uparrow$ & PPL $\downarrow$ & ROUGE-L $\uparrow$
 & Succ $\uparrow$ & PPL $\uparrow$ & ROUGE-L $\downarrow$
 & Succ $\uparrow$ & PPL $\downarrow$ & ROUGE-L $\uparrow$ & Succ.$\uparrow$\\
 
\midrule
Vanilla Model & 0.00 & 1.00 & 100.0 & 100.0 & 1.00 & 100.0 & 0.00 & 1.00 & 100.0 & 100.0 & 1.00 & 100.0 & 50.00 \\

\midrule
Gradient Ascent & 99.36 & $>10^{10}$ & 0.00 & 0.09 & $>10^{10}$ & 0.00 & 99.89 & $>10^{10}$ & 2.38 & 0.09 & $>10^{10}$ & 0.00 & 49.86\\
Fine-tuning with Random Labels & 98.00 & $10^{5}$ & 0.00 & 2.08 & $10^{5}$ & 0.00 & 99.87 & $10^{5}$ & 0.00 & 0.31 & $10^{5}$ & 0.00 & 50.07\\
Unlearning with Adversarial samples & 55.12 & 12.99 & 41.67 & 43.00 & 14.75 & 43.75 & 51.05 & 11.90 & 40.00 & 65.88 & 5.80 & 55.61 & 53.76\\

\midrule
Gradient Ascent + Descent &  &  &  &  &  &  &  &  &  &  &  &  &\\
- Descent on in-distribution data & 97.34 & $>10^{10}$ & 0.00 & 59.62 & $10^{9}$ & 37.84 & 99.79 & $>10^{10}$ & 0.00 & 77.74 & $10^{8}$ & 88.78 & 83.62 \\
- Descent on out-distribution data & 96.75 & $>10^{10}$ & 0.00 & 2.41 & $>10^{10}$ & 0.00 & 97.31 & $>10^{10}$ & 0.00 & 3.29 & $>10^{10}$ & 0.00 & 49.94\\
\midrule

Gradient Ascent + Descent &  &  &  &  &  &  &  &  &  &  &  &  &\\
- KL on in-distribution data & 99.94 & $>10^{10}$ & 0.00 & 0.88 & $>10^{10}$ & 0.00 & 100.0 & $>10^{10}$ & 0.00 & 76.74 & $10^{8}$ & 85.28 & 69.39\\
- KL on out-distribution data & 99.10 & $>10^{10}$ & 0.00 & 0.30 & $>10^{10}$ & 0.00 & 99.65 & $>10^{10}$ & 0.00 & 0.57 & $>10^{10}$ & 2.00 & 49.01\\
\midrule

Memflex & 94.40 & $>10^{10}$ & 0.00 & 72.79 & $>10^{6}$ & 75.68 & 98.15 & $>10^{10}$ & 2.38 & 89.00 & 2.49 & 91.46 & 88.59 \\
\midrule

GRAIL (Ours) & 90.72 & $>10^{10}$ & 11.11 & \textbf{85.34} & \textbf{58.81} & \textbf{94.74} & 98.75 & $>10^{10}$ & 2.38 & \textbf{93.87} & \textbf{2.44} & \textbf{97.98} & \textbf{92.17}\\
\bottomrule


        
\end{tabular}
}
\end{table*}


%% file: table_llama_general.tex
\begin{table}[!t]
\caption{General task performance Experiments on LLAMA-2-7B-Chat after unlearning} 

\vspace{-0.15cm}

\label{table:llama_general}
\resizebox{\columnwidth}{!}{
\begin{tabular}{l|c|c|c|c|c|c}
\toprule

\multirow{2}{*}{Models} & \multicolumn{6}{c}{General Task Performance} \\
\addlinespace[3pt]
\cline{2-7}
\addlinespace[3pt]
& MMLU & ARC & TruthfulQA & SIQA & RACE & Avg.\\

\addlinespace[3pt]
 
\midrule
Vanilla Model & 0.4443 & 0.6115 & 0.2913 & 0.4057 & 0.4355 & 0.4377 \\

\midrule
Gradient Ascent & 0.2295 & 0.2647 & 0.2472 & 0.2316 & 0.3429 & 0.2632 \\
Fine-tuning with Random Labels & 0.2569 & 0.2673 & 0.2264 & 0.2344 & 0.3495 & 0.2669 \\
Unlearning with adversarial samples & 0.4304 & 0.6982 & 0.2754 & 0.4105 & 0.4534 & 0.4536 \\
\midrule
Gradient Ascent + Descent &  &  &  &  &  & \\
- Descent on in-distribution data & 0.4370 & 0.4710 & 0.2656 & 0.2431 & 0.3403 & 0.3514\\
- Descent on out-distribution data & 0.4255 & 0.5968 & 0.2399 & 0.3818 & 0.3393 & 0.3967 \\
\midrule

Gradient Ascent + Descent &  &  &  &  &  &\\
- KL on in-distribution data & 0.4209 & 0.3981 & 0.2852 & 0.2574 & 0.3475 & 0.3418\\
- KL on out-distribution data & 0.4395 & 0.5328 & 0.2619 & 0.3761 & 0.3275 & 0.3876\\
\midrule

Memflex & 0.4454 & 0.5105 & 0.3231 & 0.3081 & 0.3321 & 0.3839 \\
\midrule

GRAIL (Ours) & \textbf{0.4476} & 0.6322 & 0.3023 & 0.3933 & 0.3245 & 0.4200 \\
\bottomrule
\end{tabular}
}
\end{table}
\vspace{-0.2cm}

%% file: table_qwen_unlearn_one.tex
\begin{table}[!t]
\caption{Unlearning Accuracy of Qwen-1.5-7B-Chat on User Privacy and Copyright}
\label{table:qwen_onecolumn}
\normalsize
\resizebox{\columnwidth}{!}{
\begin{tabular}{l|c|c|c|c|c}
\toprule
\multirow{2}{*}{Models} & \multicolumn{2}{c|}{Privacy} & \multicolumn{2}{c|}{Copyright} &\multirow{2}{*}{Avg.} \\
\addlinespace[3pt]
\cline{2-5}
\addlinespace[3pt]
& \multicolumn{1}{c|}{US} & \multicolumn{1}{c|}{RS} & \multicolumn{1}{c|}{US}& \multicolumn{1}{c|}{RS} \\
\addlinespace[3pt]


 
\midrule
Vanilla Model & 0.00 & 100.0 & 0.00 & 100.0 & 50.0\\

\midrule
Gradient Ascent & 99.21 & 0.11 & 99.86 & 0.10 & 49.82\\
Fine-tuning with Random Labels & 100.0 & 0.00 & 99.99 & 0.85 & 50.21\\
Unlearning with adversarial samples & 56.91 & 46.02 & 56.73 & 65.86 & 56.38\\

\midrule
Gradient Ascent + Descent  & & & & \\
- Descent on in-distribution data & 98.20 & 56.78 & 99.94 & 77.33 & 83.06\\
- Descent on out-distribution data & 100.0 & 0.00 & 100.0 & 0.00 & 50.00\\
\midrule

Gradient Ascent + Descent  & & & & \\
- KL on in-distribution data & 99.93 & 1.59 & 99.68 & 70.81 & 68.00\\
- KL on out-distribution data & 100.0 & 0.00 & 100.0 & 0.00 & 50.00\\
\midrule
Memflex & 92.02 & 71.69 & 99.07 & 81.75 & 86.13\\
\midrule

GRAIL (Ours) & 93.43 & \textbf{79.25} & 98.76 & \textbf{90.68} & \textbf{90.53}\\
\bottomrule
        
\end{tabular}
}
\end{table}

%% file: table_squential.tex
\begin{table}[!t]
\caption{Difference Unlearning Strategies for Privacy and Copyright in Llama-2-7b-Chat} \vspace{-0.15cm}
\label{table3:sequential_unlearning}
\normalsize
\resizebox{\columnwidth}{!}{
\begin{tabular}{l|c|c|c|c}
\toprule

\addlinespace[3pt]
Models & US\_{Priv} $\uparrow$ & RS\_{Priv} $\uparrow$ & US\_{Cpr} $\uparrow$ & RS\_{Cpr} $\uparrow$\\

\midrule
Vanilla Model & 0.00 & 100.0 & 0.00 & 100.0 \\
\midrule
Unlearn (Only Privacy) & 89.90 & 74.68 & 26.83(-26.83\%) & 92.77(-7.23\%) \\
Seq. Unlearn (\( P \to C \)) & 95.94 & 66.94 & 99.86 & 75.57\\
\midrule
Unlearn (Only Copyright) & 19.68(-19.68\%) & 84.96(-15.04\%) & 99.81 & 79.31 \\
Seq. Unlearn (\( C \to P \)) & 96.46 & 47.34 & 100.0 & 81.46 \\
\midrule
Unlearn Combined & 94.32 & 60.50 & 82.59 & 78.63 \\
\midrule
GRAIL (Ours) & 90.72 & \textbf{85.34} & 98.75 & 93.87 \\
\bottomrule
\end{tabular}
}

\end{table}


%% file: table_harmonic_acc.tex
\begin{table}[!t]
\caption{Harmonic success of LLAMA-2-7B-Chat after unlearning} \vspace{-0.15cm}
\label{table:HS}
\normalsize
\resizebox{\columnwidth}{!}{
\begin{tabular}{l|c|c|c}
\toprule

\multirow{2}{*}{Models} & \multicolumn{3}{c}{Harmonic Success} \\
\addlinespace[3pt]
\cline{2-4}
\addlinespace[3pt]
& Privacy & Copyright & Avg.\\

\addlinespace[3pt]
 
\midrule
Vanilla Model & 0.00 & 0.00 & 0.00 \\

\midrule
Gradient Ascent & 0.18 & 0.17 & 0.17 \\
Fine-tuning with Random Labels & 4.07 & 0.61 & 2.34 \\
Unlearning with adversarial samples & 48.31 & 57.53 & 52.92 \\
\midrule
Gradient Ascent + Descent &  &  &  \\
- Descent on in-distribution data & 73.95 & 87.4 & 80.68\\
- Descent on out-distribution data & 4.70 & 6.36 & 5.53\\
\midrule

Gradient Ascent + Descent &  &  &  \\
- KL on in-distribution data & 1.74 & 86.84 & 44.29\\
- KL on out-distribution data & 0.61 & 1.14 & 0.88 \\
\midrule

Memflex & 82.20 & 93.35 & 87.78 \\
\midrule

GRAIL (Ours) & \textbf{87.95} & \textbf{96.25} & \textbf{92.10} \\
\bottomrule
\end{tabular}
}
\end{table}

%% file: table_ablation_GRAIL.tex
\begin{table}[!t]
\caption{Ablation study of GRAIL on LLaMA-2-7B-Chat}
\label{table:ablation_GRAIL}
\normalsize
\resizebox{\columnwidth}{!}{
\begin{tabular}{l|c|c|c|c|c}
\toprule
\multirow{2}{*}{Models} & \multicolumn{2}{c|}{Privacy Success} & \multicolumn{2}{c|}{Copyright Success} &\multirow{2}{*}{Avg.} \\
\addlinespace[3pt]
\cline{2-5}
\addlinespace[3pt]
& \multicolumn{1}{c|}{Unlearning} & \multicolumn{1}{c|}{Retention} & \multicolumn{1}{c|}{Unlearning}& \multicolumn{1}{c|}{Retention} \\
\addlinespace[3pt]

\midrule
GRAIL & 90.72(-3.66\%) & \textbf{85.34(+13.51\%)} & 98.75(-1.20\%) & \textbf{93.87(+20.22\%)} & \textbf{92.17} \\
\midrule
w/o OP-UR & 94.62(+0.48\%) & 72.84(-3.11\%) & 98.77(-1.18\%) & 89.70(+14.88\%) & 90.17 \\
w/o OP-RR & 91.11(-3.25\%) & 78.68(+4.66\%) & 96.50(-3.45\%) & 92.23(+18.12\%) & 89.63 \\
w/o both & 94.17(+0.00\%) & 75.18(+0.00\%) & 99.95(+0.00\%) & 78.08(+0.00\%) & 86.85\\
\bottomrule

\end{tabular}
}
\end{table}